\documentclass{article}
\usepackage{ijcai22}

\usepackage{times}
\usepackage{soul}
\usepackage{url}
\usepackage[hidelinks]{hyperref}
\usepackage[utf8]{inputenc}
\usepackage[small]{caption}
\usepackage{graphicx}
\usepackage{amsthm}
\usepackage{booktabs}
\usepackage{algorithm,algorithmicx,algpseudocode}
\usepackage{amsmath,amsfonts,bm}
\usepackage{enumitem}
\usepackage{cancel}
\usepackage{amsmath,amsfonts,bm}

\def\eqref#1{equation~\ref{#1}}

\def\1{\bm{1}}

\DeclareMathAlphabet{\mathsfit}{\encodingdefault}{\sfdefault}{m}{sl}
\SetMathAlphabet{\mathsfit}{bold}{\encodingdefault}{\sfdefault}{bx}{n}

\DeclareMathOperator*{\argmax}{arg\,max}

\floatstyle{ruled}
\newfloat{model}{thp}{lop}
\floatname{model}{Model}

\def\code#1{\texttt{#1}}
\DeclareMathOperator*{\Minimize}{Minimize}
\usepackage{hyperref}
\usepackage{color}
\usepackage[colorinlistoftodos]{todonotes}

\newcommand{\tlf}{{\sc 2SL-FJSP}}

\title{Two-Stage Learning For the Flexible Job Shop Scheduling Problem}

\author{
    W
    \affiliations
    Affiliation
    \emails
    pcchair@ijcai-22.org
}

\author{
Wenbo~Chen
\and
Reem~Khir
\and
Pascal~Van~Hentenryck
\affiliations
Georgia Institute of Technology, Atlanta, USA\\
\emails
\{wenbo.chen, reem.khir\}@gatech.edu,
pascal.vanhentenryck@isye.gatech.edu,
}

\begin{document}

\maketitle

\begin{abstract}
The Flexible Job-shop Scheduling Problem (FJSP) is an important
combinatorial optimization problem that arises in manufacturing and
service settings. FJSP is composed of two subproblems, an assignment
problem that assigns tasks to machines, and a scheduling problem that
determines the starting times of tasks on their chosen machines.
Solving FJSP instances of realistic size and composition is an ongoing
challenge even under simplified, deterministic assumptions. Motivated
by the inevitable randomness and uncertainties in supply chains,
manufacturing, and service operations, this paper investigates the
potential of using a deep learning framework to generate fast and
accurate approximations for FJSP. In particular, this paper proposes
a \textit{two-stage} learning framework \tlf{} that explicitly models
the hierarchical nature of FJSP decisions, uses a confidence-aware
branching scheme to generate appropriate instances for the scheduling
stage from the assignment predictions, and leverages a novel
symmetry-breaking formulation to improve learnability. \tlf{} is
evaluated on instances from the FJSP benchmark library. Results show
that \tlf{} can generate high-quality solutions in milliseconds,
outperforming a state-of-the-art reinforcement learning approach
recently proposed in the literature, and other heuristics commonly
used in practice.
\end{abstract}

\section{Introduction}

The Flexible Job-shop Scheduling problem (FJSP) is a generalization of
the classical Job-shop Scheduling problem (JSP) that appears in a
wide-range of application areas including logistics, manufacturing,
transportation, and services. FJSP considers a set of jobs, each
consisting of a number of ordered tasks that need to be assigned and
scheduled on machines such that the maximum time needed to complete
all jobs, known as the makespan, is minimized. The FJSP is hard to
solve for real-sized instances, even under simplified, deterministic
assumptions; it is NP-hard in the strong sense.

The FJSP is composed of two subproblems, an assignment subproblem that
assigns tasks to machines, and a scheduling subproblem that determines
the sequence and processing schedule of tasks on their selected
machines \cite{brandimarte1993routing}. For a complex problem like the
FJSP, \textit{concurrent} approaches that solve for the assignment and
scheduling subproblems simultaneously often fail to provide effective
solutions in reasonable
runtimes \cite{chaudhry2016research}. \textit{Hierarchical}
approaches, on the other hand, are more common as they reduce the
original problem complexity by decomposing it into subproblems that
are solved sequentially. A natural hierarchical scheme for
solving the FJSP is based on the observation that, when machine
assignments are decided, the FJSP reduces to the classical JSP. For
that reason, \textit{two-stage} hierarchical schemes that separate
assignment and scheduling decisions have been widely used in the design
of efficient heuristic and exact FJSP algorithms (see for
example: \cite{fattahi2007mathematical,xie2019review,naderi2022critical},
and \cite{xie2019review} for a comprehensive review).

While extensive research has been done to solve FJSP using exact and
approximate solution techniques, operational realities surrounding
its application domain hinder their effective adoption especially in
contexts when solutions must be generated under stringent time
constraints, \textit{i.e.} in real time. For example, uncertainty
revealed on day-of-operation is a key practical concern that may
require immediate replanning of operations. Such uncertainty could
stem from either the continuous variations of processing times,
and/or the combinatorial presence, or absence, of machines, or jobs
for processing. This work focuses on the former aspect that if not
addressed properly, could result in machine idling or job delays that
can be very costly. To address this issue, planners typically resort
to using rule-based, hierarchical, heuristics that can generate
approximate solutions very
quickly \cite{jun2019learning,kawai2005efficient}. However, while
efficient, the sub-optimality of their solutions could translate into
considerable economical losses that are not desirable in practice.

Motivated by the need for fast and effective approximate solutions,
the training of machine-learning models as surrogates for solving hard
combinatorial optimization problems has gained traction;
see \cite{vesselinova2020learning} for a survey on the topic. An
important challenge in this line of work is related to ensuring the
feasibility of predicted solutions. To address this, several
approaches have been proposed in the literature including the use of
Langrangian loss functions (see for
example \cite{kotary2022fast,fioretto2020predicting,chatzos2020high}),
and incorporating optimization algorithms into differentiable systems
(see for
example \cite{vinyals2015pointer,khalil2017learning,kool2018attention,park2022confidence}). Another
challenge is related to the solution structure that is present in
combinatorial problems. For a problem like FJSP, it is common to have
multiple, symmetric, solutions since there can be multiple identical
machines that can process the same
job \cite{ostrowski2010symmetry}. Additionally, it is also common to
work with solutions that are approximate and not optimal due to the
computationally difficulty associated with solving instances to
optimality. The presence of these effects could result in datasets
that are harder to learn, and therefore, more structured data
generation approaches are needed \cite{kotary2021learning}.

To this end, this paper proposes \tlf{}, a machine-learning framework
that can be used as an optimization proxy for solving FJSP instances
in uncertain environments. \tlf{} exploits the problem structure and
proposes a \textit{two-stage} deep-learning approach that learn to map
FJSP instances, sampled from a distribution of processing times, to
assignment and scheduling solutions that are close to those generated
directly via optimization. The \textit{two-stage} structure of \tlf{}
is complemented by a confidence-aware branching scheme that generates
instances for the second-stage learning to account for prediction
errors in the first stage. \tlf{} also leverages a novel data
generation that exploits a novel symmetry-breaking scheme that
may facilitate learning significantly. 

\paragraph{Contributions}
The contributions of this paper can be summarized as follows. (1) It
proposes \tlf{}, a \textit{two-stage} learning framework for the FJSP
that exploits its hierarchical nature in providing fast
and effective FJSP approximations. (2) \tlf{} introduces a confidence-aware branching scheme to account for prediction errors in the assignment stage and
generate a more comprehensive set of instances for the scheduling stage. (3)
\tlf{} includes a novel symmetry-breaking data generation approach
that alleviates issues in learnability caused by the presence of
co-optimal and approximate solutions typically experienced in
scheduling problems. The symmetry-breaking scheme breaks the
symmetries for all instances uniformly and can be parallelized
contrary to prior work. (4) \tlf{} integrates these innovations with
state-of-the-art techniques for capturing constraint violations and
restoring feasibility. (5) \tlf{} has been evaluated on standard
benchmarks and computational results show that, on the experimental
setting considered in the paper, it significantly outperforms
state-of-the-art heuristic and reinforcement learning approaches. The results
also show the critical role of each component.

\section{Related Work}

\paragraph{Learning for constrained optimization}

Deep learning is increasingly being applied to constrained
optimization problems.  The typical approaches use neural networks to
predict permutations or combinations as sequences of actions by using
imitation learning or reinforcement
learning \cite{vinyals2015pointer,khalil2017learning,kool2018attention,park2022confidence}.
Other approaches involve using (unrolled) optimization algorithms as
differentiable layers in neural
networks \cite{amos2017optnet,agrawal2019differentiable,vlastelica2019differentiation,chen2020rna,donti2021dc3}.
Additionally, some works regularize the training of the neural network
with penalty methods or Lagrangian
duality \cite{fioretto2020predicting,kotary2021learning,chatzos2020high}.
Refer to \cite{bengio2021machine,kotary2021end,lei2022multi} for more
comprehensive reviews.

\paragraph{Learning for Flexible Job Shop Scheduling}

Recently, a number of works use deep reinforcement learning (DRL) for
Job Shop Scheduling Problems
(JSP) \cite{lin2019smart,park2019reinforcement,zhang2020learning,song2022flexible,liu2022deep}.
In \cite{lin2019smart}, a deep Q learning-based method is used to
select the priority dispatch rule of JSP.
\cite{zhang2020learning} represents the disjunctive graph of JSP using graph neural networks to embed the state information and train the RL agent using proximal policy optimization \cite{schulman2017proximal}.
\cite{song2022flexible} extends the work \cite{zhang2020learning} to FJSP using heterogeneous graph representation.
However, previous DRL works encounter convergence issues when training
on large instances. In particular, as shown
in \cite{song2022flexible}, increasing the instance size in the DRL
training often degrades the performance of the DRL policy on
public benchmarks.

Recently, \cite{kotary2022fast} explored another direction to learn
the mapping from the processing times to optimal schedules by using
Lagrangian duality for the job-shop scheduling problem. \tlf{} builds
on this work and proposes a two-stage learning framework to
approximate the FJSP. \tlf{} explicitly models the hierarchical nature
of the FJSP decisions, uses a confidence-aware neural network for the
assignment predictions to generate training instances for the scheduling
predictions, and proposes a new symmetry-breaking scheme to generate,
in parallel, the training data for the scheduling predictions.

\section{Preliminaries}

\paragraph{The Flexible Job-shop Scheduling Problem}

The Flexible Job-shop Scheduling Problem (FJSP) is defined primarily
by a set of jobs $J$ and a set of machines $M$. Each job consists of a
set of ordered tasks $T$ that need to be processed in a
pre-defined sequence. Each task is allowed to be processed on one
machine out of a set of alternative machines. The
expected duration of task $t$ of job $j$ on machine $m$ is denoted by $d^m_{jt}$
and $N = J \times T$ is the set of all tasks.

Model 1 provides a high-level mathematical description of the FJSP
with decision variables $z^m_{jt} \in \{1,0\}$ indicating whether task
$t$ of job $j$ is assigned to machine $m$ for processing,
$s^m_{jt} \geq 0$ representing the start time of task $t$ of job $j$
on machine $m$, and $C_{max} \geq 0$ representing the maximum
completion time of all jobs, also known as the makespan. The FJSP aims
at finding task-to-machine assignments and their corresponding
schedules such that the makespan is minimized.

The \textit{assignment} constraints (\ref{fjsp_assignment}) ensure
that a task is processed on a single machine without
splitting. The \textit{task-precedence} constraints
(\ref{fjsp_taskPrecedence}) ensure that all tasks are processed
following their given order; for simplicity, a higher-indexed task is
a predecessor of a lower-indexed task. The \textit{machine-precedence}
constraints (\ref{fjsp_machinePrecedence}) ensure that no two tasks
overlap in time if they are assigned to be processed on the same
machine. The disjunctive nature of constraints
(\ref{fjsp_machinePrecedence}) makes the FJSP hard to solve even for
medium-sized instances. It can be modeled in various ways using e.g.,
Mixed Integer Programming (MIP) and Constraint Programming (CP), each
having its own features and merits
(e.g., \cite{chaudhry2016research,fattahi2007mathematical}). This paper uses
a CP approach that is proven to be effective for solving scheduling
problems; the corresponding FJSP CP formulation is presented in 
Appendix A. 








\begin{model}[ht!]
\begin{subequations}
{\scriptsize
\caption{FJSP Description}
\label{model:fjsp_description}
\begin{flalign} 
&\mbox{Minimize} \quad C_{max} \qquad \label{eq:fjsp_obj_simple}\\
&\mbox{subject to:}\nonumber\\
&\sum_{m \in M_{jt}} z^m_{jt} = 1  \qquad \forall j \in [J], t \in [T] \label{fjsp_assignment}
\\
&s^m_{jt-1} \geq  s^m_{jt} + d^m_{jt}z^m_{jt}  \qquad  \forall j \in [J], t \in [T], t>0, m \in M_{jt},\label{fjsp_taskPrecedence}\\ 
 & s^m_{jt} \geq  s^m_{j^\prime t^\prime} + d^m_{j^\prime t^\prime}z^m_{j^\prime t^\prime}  \quad \vee \quad s^m_{j^\prime t^\prime} \geq  s^m_{jt} + d^m_{jt}z^m_{jt},   \label{fjsp_machinePrecedence}\\
&  \qquad \qquad \forall j, j^\prime \in [J], t,t^\prime \in [T], j\neq j^\prime, m \in M_{jt},
 \nonumber \\ 
& C_{max} \geq s^m_{jt} + d^m_{jt}z^m_{jt}  \qquad  \forall j \in [J], t \in [T], t=0, m \in M_{jt}, \label{fjsp_ObjLinearization}  \\ 
& z^m_{jt} \in \{0,1\} \qquad  \forall j \in [J], t \in [T], m \in M_{jt}, \\
& s^m_{jt} \geq 0 \qquad  \forall  j \in [J], t \in [T], m \in M_{jt}.
\end{flalign}
}
\end{subequations}
\end{model}

\paragraph{The Job Shop Scheduling Problem}

The Job Shop Scheduling (JSP) is a special case of the FJSP, with each
task having a single machine option for processing. This reduces the
decision space to finding task schedule given a known
assignment. Model 2 presents a high-level description of the JSP with
decision variables $s_{jt}$ and $C_{max}$ representing the start time
of each task and the makespan, respectively. The description of
the \textit{task-precedence} constraints (\ref{jsp_taskPrecedence})
and the \textit{machine-precedence} constraints
(\ref{jsp_machinePrecedence}) is identical to that of FJSP;
no \textit{machine selection} is needed in JSP since a task can be
processed only on a specified machine.



\begin{model}[ht!]
\begin{subequations}
{\scriptsize
\caption{JSP Description}
\label{model:jsp_description}
\begin{flalign} 
&\mbox{Minimize} \quad C_{max} \qquad \label{eq:jsp_obj_simple}\\
&\mbox{subject to:}\nonumber\\
&s_{jt-1} \geq  s_{jt} + d_{jt}  \qquad  \forall j \in [J], t \in [T], t>0,\label{jsp_taskPrecedence}\\ 
& s_{jt} \geq  s_{j^\prime t^\prime} + d_{j^\prime t^\prime}  \quad \vee \quad s_{j^\prime t^\prime} \geq  s_{jt} + d_{jt}   \qquad  \label{jsp_machinePrecedence}\\
&  \qquad \forall j, j^\prime \in [J], t, t^\prime \in [T],  j\neq j^\prime, \sigma_{jt}=\sigma_{j^\prime t^\prime}
 \nonumber \\ 
& C_{max} \geq s_{jt} + d_{jt}  \qquad \forall j \in [J], t \in [T], t=0 \label{jsp_objLinearization} \\ 
& s_{jt} \geq 0 \qquad \forall j \in [J], t \in [T], m \in M_{jt}.
\end{flalign}
}
\end{subequations}
\end{model}

\section{The Learning Task for the FJSP}

Following \cite{kotary2022fast}, the learning task is motivated by a
real-time operational setting where some machine experience an
unexpected ``slowdown'', resulting in increased processing times for
each task assigned to the impacted machine. The goal of the
machine-learning model is to quickly find a new solution to the
resulting FJSP. The instances are thus specified by the set of
processing times for each machine and each task compatible set of
machines. Formally, the goal of FJSP learning task is to construct a
parametric model $f_{\boldsymbol{\theta}}: \mathbb{N}^{N\times
M} \rightarrow \{0,1\}^{N \times M} \times \mathbb{N}^{N}$ which,
given the durations of the tasks on the machines, predicts the optimal
machine assignments $\mathbf{z} \in \{0,1\}^{N \times M}$ and the
optimal start times $\mathbf{s} \in \mathbb{N}^{N}$ for each task.
Such a parametric model is often called an optimization proxy or an
optimization surrogate.

Consider dataset $\mathcal{D}= \{(\mathbf{d}^i, \mathbf{z}^i, \mathbf{s}^i)\}_{i=1}^D$ with $D$
instances, where the set of processing times is denoted by
$\mathbf{d} \in \mathbb{N}^{N\times M}$, $N$ is the total number of
tasks $N = J*T$, and the processing time $d^{n,m}$ is $\infty$ when
task $n$ is not compatiable with machine $m$. The learning task
corresponds to solving the following optimization problem:
\begin{align}
    \min_{\boldsymbol{\theta}} \sum_i^D \mathcal{L}(f_\theta(\mathbf{d}^i), \mathbf{s}^i, \mathbf{z}^i) \\
    \text{subject to:} \quad \mathcal{C}(\mathbf{d}^i, f_\theta(\mathbf{d}^i)),
\end{align}
where $\mathcal{C}(\mathbf{d}^i, f_\theta(\mathbf{d}^i))$ holds if the
predicted assignment and start time satisfy the constraints of
Model \ref{model:fjsp_description}. This learning task
is challenging for the following reasons:
\begin{itemize}

\item[1.] The decision space of the FJSP has a hierarchical
    nature. Not capturing this structure results in approximations
    with high optimality gaps (as shown in the experiments).

\item[2.] The FJSP has highly combinatorial feasibility constraints. Traditional
    parametric models typically fail to capture such constraints.

\item[3.] The FJSP is highly combinatorial, and the optimal solutions have many
    symmetries, which raises significant challenges for machine learning.
\end{itemize}

\section{A Two-Stage Learning Framework}

To address these challenges, the paper proposes \tlf{}, a learning
framework for the FJSP based on three novel contributions that exploit the
underlying structure of the problem:
\begin{itemize}

\item[1.] \tlf{} is a {\em two-stage} learning where the assignment decisions are learned
first, before learning the starting times;

\item[2.] \tlf{} uses a {\em confidence-aware branching} scheme to generate 
appropriate instances to train the second-stage learning models;

\item[3.] \tlf{} uses a novel {\em symmetry-breaking} formulation of FJSP to bias the CP solver
and facilitate the learning task. The formulation makes it possible to solve all instances
in parallel contrary to earlier work where the data generation process is sequential.
\end{itemize}

\begin{figure*}[!t]
\begin{center}
\includegraphics[width=1.\textwidth]{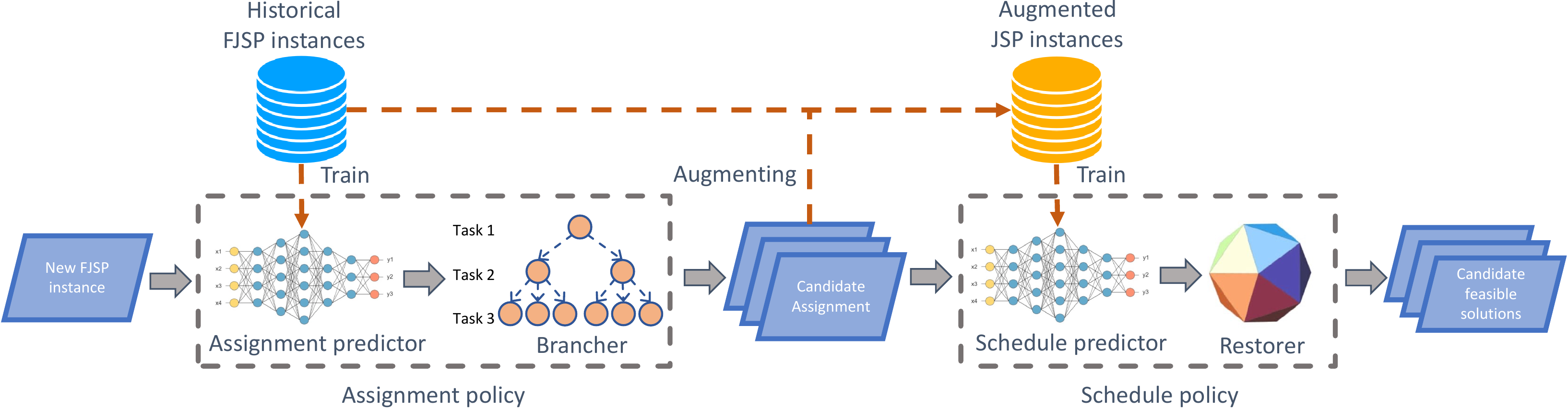}
\caption{Two-stage Learning Framework for FJSP (2SL-FJSP)}
\label{img:FJSPLearning:pipeline}
\end{center}
\end{figure*}

\noindent
Figure \ref{img:FJSPLearning:pipeline} presents \tlf{}, the two-stage
learning framework for the FJSP. Its various components and their
contributions are explained in detail in this section.  At a high
level, \tlf{} sequentializes the learning of assignment and scheduling
decisions. Importantly, the first-stage assignment predictions are
used through a branching process to generate instances used in the
second-stage learning task, i.e., the predictions of the start times.


\subsection{The Neural Network Structure}
\label{sec: architecture}

The parametric models used by \tlf{} for predicting machine
assignments and start times are inspired by the design proposed
in \cite{kotary2021learning}. The raw processing times $\mathbf{d}$ is
split into two matrices: machine-task matrix
$\mathbf{d}_m \in \mathbb{N}^{N\times M}$ and job-task matrix
$\mathbf{d}_j \in \mathbb{N}^{N\times J}$, representing the
machine-task structure and the job-task structure in FJSP. A distinct
encoder is used on each matrix. The two encodings are then
concatenated to form the inputs of a decoder that predicts the machine
assignments or the start times.
\begin{align*}
    & \mathbf{h}_m = \text{Enc}_{\mathbf{\boldsymbol{\theta}}_1}(\mathbf{d}_m), \\  
    & \mathbf{h}_j = \text{Enc}_{\mathbf{\boldsymbol{\theta}}_2}(\mathbf{d}_j), \\
    & \hat{\mathbf{y}} = \text{Dec}_{\mathbf{\boldsymbol{\theta}}_3}([\mathbf{h}_j; \mathbf{h}_2]), 
\end{align*}

\noindent
The encoders
$\text{Enc}_{\mathbf{\boldsymbol{\theta}}_1}: \mathbb{N}^{N\times
M} \rightarrow \mathbb{N}^{N\times H}$ and
$\text{Enc}_{\mathbf{\boldsymbol{\theta}}_2}: \mathbb{N}^{N\times
J} \rightarrow \mathbb{N}^{N\times H}$ are stacked 1-D convolution
layers with $H$ filters.  The decoder
$\text{Dec}_{\mathbf{\boldsymbol{\theta}}_3}$ is a multi-layer
perceptron that maps the flatted latent representation to the optimal
decisions. 

\subsection{The Confidence-Aware Branching Scheme}

The assignment predictions are not necessarily correct, which raises
an interesting issue: {\em the dataset of instances may not contain an
instance with this particular (predicted) assignment}. \tlf{}
addresses this challenge by using a confidence-aware branching scheme.
The first element of the branching scheme is the parametruc model for
the assignment predictions that is trained in a supervised learning
fashion by minimizing the cross-entropy loss. {\em The outputs
$\hat{\mathbf{z}}$ for the machine-assignment layer are logits and
they play a critical role to interface the assignment and scheduling
policies.} Given the logits $\mathbf{\hat{z}} \in \mathbb{R}^{N\times
M}$ predicted by the deep learning model, the probability of a task
$n$ being assigned to the machine $m$ is computed as:
\begin{align}
p_{nm} = \frac{\exp(\hat{z}_{nm})\mathbb{I}\{d^{nm} \neq \infty\}}{\sum_{m'\in[M]}\exp(\hat{z}_{nm'})\mathbb{I}\{d^{nm'} \neq \infty\}},
\end{align}
where $d^{nm} \neq \infty$ indicates task $n$ is compatible with
machine $m$. The loss function $\mathcal{L}_c(\mathbf{p}, \mathbf{z})$
of the first-stage predictive model is then defined as
\begin{align}
\label{eq:cls:loss}
& -\sum_{\substack{n, m}} z_{nm}\log {p}_{nm} +(1-z_{nm}) \log (1-p_{nm})
\end{align}

\noindent
To generate appropriate instances for training the scheduling policy,
\tlf{} uses {\em confidence-aware branching}, a strategy to generate a set of candidate
assignments based on the confidence of the predictions. The candidate
solutions are then used to generate the training data set for the
scheduling policy. The confidence of the DNN prediction is based on
the softmax function. More advanced uncertainty quantification such as
MCDropout \cite{gal2016dropout} and deep
ensembling \cite{rahaman2021uncertainty} could be also used.

Confidence-aware branching borrows ideas from neighborhood search.  It
generates new candidate solutions by fixing the high-confidence
predictions and perturbing the assignment predictions with lower
confidence levels. More precisely, confidence-aware branching selects
the subset $\mathcal{N}_s$ of tasks with the highest entropy, i.e.,
\begin{equation}
\mathcal{N}_s = \argmax_{\mathcal{N}_s \in [N]} \sum_{n \in \mathcal{N}_s} \sum_{m \in [M]} -p_{nm}\log p_{nm},
\end{equation}
and branches on all compatible assignments for every task
$n \in \mathcal{N}_s$.  The generated candidate assignments are used
to augment the dataset for training the scheduling policies.

\subsection{The Scheduling Policy}

The scheduling policy of \tlf{} is trained to imitate the CP
solver. The loss function for its neural network minimizes a weighted
sum of prediction errors and constraint violations, i.e.,
\begin{align}
    \mathcal{L}_r(\mathbf{s}, \hat{\mathbf{s}}, \mathbf{z}, \mathbf{d}) = \mathcal{L}(\mathbf{s}, \hat{\mathbf{s}}) + \lambda \sum_{c\in \mathcal{C}}v_c(\hat{\mathbf{s}}, \mathbf{z}, \mathbf{d}),
\end{align}
where $\hat{\mathbf{s}}$ denotes the predicted starting times, the
function $\mathcal{L}(\mathbf{s}, \hat{\mathbf{s}}) = \|\mathbf{s}
- \hat{\mathbf{s}}\|_1$ measures the distance of the predicted
starting times with the optimal starting times,
$v_c(\hat{\mathbf{s}}, \mathbf{z}, \mathbf{d})$ denotes the constraint
violation of constraint $c$, and $\lambda$ denotes the coefficient of
the constraint regularizer. Given the machine assignment $\mathbf{z}$
and the predicted starting times $\hat{\mathbf{s}}$, the job precedence
constraint violation is computed as:
\begin{align}
    v_{1c}(\hat{s}_{jt}) & = \max(0, \hat{s}_{jt} + d_{jt}^m - \hat{s}_{jt-1}).
\end{align}
The machine disjunctive constraint violation is given by:
\begin{equation}
    v_{1d}(\hat{s}_{jt}, \hat{s}_{j't'}) = \min(v_{1d}^L(\hat{s}_{jt}, \hat{s}_{j't'}), v_{1d}^R(\hat{s}_{jt}, \hat{s}_{j't'})),
\end{equation}
with 
\begin{align*}
& v_{1d}^L(\hat{s}_{jt}, \hat{s}_{j't'}) = \max(0, \hat{s}_{j't'}+d_{j't'}^m - \hat{s}_{jt}), \\
& v_{1d}^R(\hat{s}_{jt}, \hat{s}_{j't'}) = \max(0, \hat{s}_{jt}+d_{jt}^m - \hat{s}_{j't'})
\end{align*}
where $d^m_{ji}$ is the processing time on the assigned machine $m$ determined by the input assignment $\mathbf{z}$.

Although the constraint regularization pushes the predicted starting
times into the feasible region, the predictions often violate some of
the constraints. To restore feasibilty, the proposed learning
framework uses the feasibility recovery in \cite{kotary2021learning}
that runs in polynomial time. The details are deferred in
Appendix B. 


\subsection{Symmetry-Breaking Data Generation}

The last important contribution of \tlf{} is a symmetry-breaking data
generation that significantly facilitates the learning problem.  The
presence of symmetries in the FJSP raises additional difficulties for
the learning task. Indeed, multiple identical machines can process the
same job without affecting the
makespan \cite{ostrowski2010symmetry}. In other words, given a
feasible assignment solution $z$, multiple equivalent solutions can be
obtained by permuting the columns of $z$. In optimization,
symmetry-breaking techniques are often used to accelerate the
search. However, different optimization instances may not break
symmetry in the same way, leading to solutions that are quite
distinct. This results in datasets that are harder to learn.  In
addition, for a complex problem like FJSP, it is not uncommon to
generate sub-optimal solutions when optimal solutions cannot be
obtained before the time limit. This poses another challenge for the
learning task as different runs of the same, or similar, instances,
may result in drastically different solutions. As a result, it is
important to go beyond the \textit{standard} approach that generates
datasets by solving each instance independently.

\begin{figure}[!t]
\includegraphics[width=.48\textwidth]{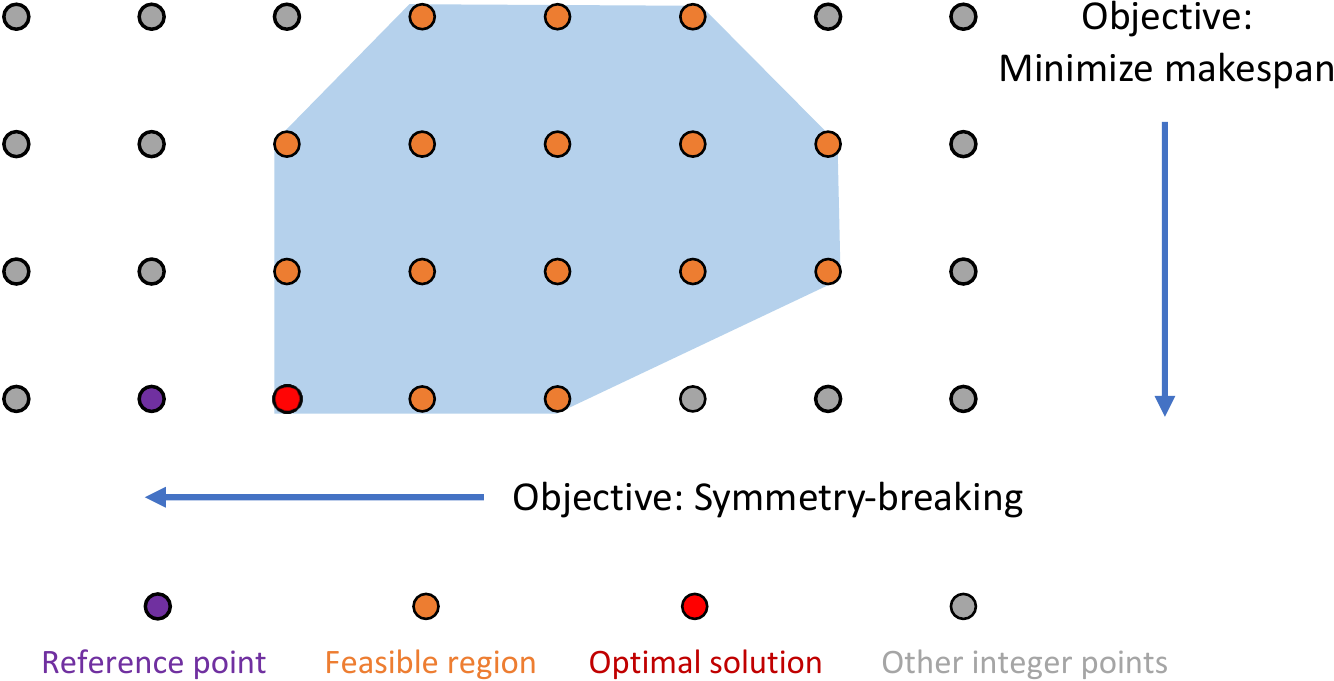}
\caption{Illustration of the Symmetry-breaking Data Generation}
\label{img:FJSPLearning:symmetry-breaking}
\end{figure}

\tlf{} features a novel symmetry-breaking data
generation approach. The scheme is inspired by the approach proposed
in \cite{kotary2021learning}, but with an important
difference. Instead of the {\em sequential} data generation framework
that minimizes variations between consecutive ordered solutions,
{\em \tlf{} uses a single reference point to break symmetries. As a
result, \tlf{} can solve all instances in parallel and all instances
break symmetries the same way.}
Figure \ref{img:FJSPLearning:symmetry-breaking} illustrates the
intuition behind the symmetry-breaking approach. In the figure, there
are three co-optimal solutions from a standard FJSP perspective. The
symmetry-breaking approach of \tlf{} directs the search towards the
solution that is closer to the pre-specified reference
solution. Model \ref{model:fjsp_SB} formalizes the symmetric-breaking
scheme of \tlf{} for the assignment policy. The idea is similar for
the scheduling policy. In the model, $\hat{\textbf{z}}$ represents the
reference assignment solution and $C_{max}^*$ is the optimal objective
of the FJSP which is obtained in a first step.  The objective
(\ref{fjsp_SB_obj}) minimizes the deviation of the optimal solution
from a reference solution using the Hamming distance. Constraints
(\ref{fjsp_SB_machineSelection})-(\ref{fjsp_SB_machinePrecedence}) are
the assignment, task-precedence, and machine constraints,
respectively; they are identical to constraints
(\ref{fjsp_taskPrecedence})-(\ref{fjsp_machinePrecedence}). Constraint
(\ref{fjsp_SB_makespan}) ensures that the quality of the
symmetry-breaking solution is not worse than the optimal solution to
the original FJSP. The experiments uses the CP formulation presented
in Appendix C 
to solve Model \ref{model:fjsp_SB}.

\begin{model}[t!]
\begin{subequations}
{\scriptsize
\caption{FJSP Symmetry Breaking Description}
\label{model:fjsp_SB}
\begin{flalign} 
&\mbox{Minimize} \quad  \sum_{j, t, m} | z^m_{jt} - \hat{z}^m_{jt}| \label{fjsp_SB_obj} \\
&\mbox{subject to:}\nonumber\\
&\sum_{m \in M_{jt}} z^m_{jt} = 1  \qquad  \forall j \in [J], t \in [T], \label{fjsp_SB_machineSelection} \\
&s^m_{jt-1} \geq  s^i_{j} + d^m_{jt}z^m_{jt}  \qquad   \forall j \in [J], t \in [T],t>0,  m \in M_{jt},\label{fjsp_SB_jobPrecedence}\\ 
& s^m_{jt} \geq  s^m_{j^\prime t^\prime} + d^m_{j^\prime t^\prime}z^m_{j^\prime t^\prime}  \quad  \vee \quad s^m_{j^\prime t^\prime} \geq  s^m_{jt} + d^m_{jt}z^i_{jt}  \label{fjsp_SB_machinePrecedence}\\
&  \qquad    \forall j, j^\prime \in [J], t, t^\prime \in [T], j\neq j^\prime, m \in M_{jt},
 \nonumber \\
& C^*_{max}  \leq  s^m_{jt} + d^m_{jt}z^m_{jt} \;  \qquad  \forall (j,t)\in T, t=0, m \in M_{jt},  \label{fjsp_SB_makespan}\\
& z^m_{jt} \in \{0,1\} \qquad  \forall j \in [J], t \in [T], m \in M_{jt} \label{fjsp_SB_z} \\
& s^m_{jt} \geq 0 \qquad   \forall j \in [J], t \in [T], m \in M_{jt}
\end{flalign}
}
\end{subequations}
\end{model}

\section{Experimental Results}

\begin{table*}[ht!]
\centering
\caption{Performances of models on FJSP instances. The best results are shown in bold. The relative optimality gap (Gap) is in percentage (\%) and the solving time (Sol.T) is reported in second. The instances have sizes of the form ($J$ $\times$ $T$).}
\label{tab: res: low_flexibility}
\resizebox{2.05\columnwidth}{!}{
\begin{tabular}{lrrrrrrrrrrrrrr}
\toprule
Method & \multicolumn{10}{c}{Low flexibility} & \multicolumn{4}{c}{High flexibility} \\
\cmidrule(lr){2-11} \cmidrule(lr){12-15} \\
 & \multicolumn{2}{c}{car8 (8$\times$8)} & \multicolumn{2}{c}{mt10 (10$\times$10)} & \multicolumn{2}{c}{la22 (15$\times$10)} & \multicolumn{2}{c}{la32 (30$\times$10)} & \multicolumn{2}{c}{la40 (15$\times$15)} & \multicolumn{2}{c}{mt10 (10$\times$10)} & \multicolumn{2}{c}{la22 (15$\times$10)} \\
 & \multicolumn{1}{c}{Gap} & \multicolumn{1}{c}{Sol. T} & \multicolumn{1}{c}{Gap} & \multicolumn{1}{c}{Sol. T} & \multicolumn{1}{c}{Gap} & \multicolumn{1}{c}{Sol. T} & \multicolumn{1}{c}{Gap} & \multicolumn{1}{c}{Sol. T} & \multicolumn{1}{c}{Gap} & \multicolumn{1}{c}{Sol. T} & \multicolumn{1}{c}{Gap} & \multicolumn{1}{c}{Sol. T} & \multicolumn{1}{c}{Gap} & \multicolumn{1}{c}{Sol. T} \\
 \midrule
CP Solver & 0.00 & 1.368 & 0.00 & 1.430 & 0.00 & 4.984 & 0.00 & 354.40 & 0.00 & 152.26 & 0.00 & 0.530 & 0.00 & 1315.0 \\
 \midrule
  2SL-FJSP& 1.47 & \textbf{0.001} & \textbf{2.70} & \textbf{0.002} & \textbf{5.40} & \textbf{0.002} & \textbf{9.50} & \textbf{0.012} & \textbf{3.16} & \textbf{0.020} & \textbf{1.84} & \textbf{0.002} & \textbf{12.38} & \textbf{0.003} \\
Encoder & \textbf{0.26} & \textbf{0.001} & 6.04 & \textbf{0.002} & 15.48 & \textbf{0.002} & 12.71 & \textbf{0.012} & 12.68 & \textbf{0.020} & 7.29 & \textbf{0.002} & 28.11 & \textbf{0.003} \\
DRL & 25.18 & 0.594 & 28.55 & 0.779 & 26.34 & 1.394 & 27.26 & 2.459 & 26.88 & 1.769 & 61.43 & 0.083 & 63.94 & 0.145 \\
 \midrule
FIFO\_SPT & 27.10 & 0.054 & 35.61 & 0.081 & 33.37 & 0.144 & 19.53 & 0.404 & 34.44 & 0.231 & 29.43 & 0.080 & 39.31 & 0.137 \\
FIFO\_EET & 19.18 & 0.049 & 26.19 & 0.079 & 26.58 & 0.139 & 16.63 & 0.383 & 26.40 & 0.226 & 60.26 & 0.093 & 61.5 & 0.167 \\
MOPNR\_SPT & 41.18 & 0.056 & 37.88 & 0.094 & 32.86 & 0.167 & 20.33 & 0.499 & 35.53 & 0.270 & 33.27 & 0.088 & 41.77 & 0.157 \\
MOPNR\_EET & 41.78 & 0.053 & 35.98 & 0.090 & 30.17 & 0.159 & 20.87 & 0.461 & 33.32 & 0.257 & 89.69 & 0.087 & 106.38 & 0.150 \\
LWKR\_SPT & 38.34 & 0.050 & 56.90 & 0.087 & 70.02 & 0.152 & 58.41 & 0.416 & 79.65 & 0.244 & 101.47 & 0.085 & 115.36 & 0.147 \\
LWKR\_EET & 39.50 & 0.049 & 56.56 & 0.086 & 64.3 & 0.149 & 61.11 & 0.406 & 90.92 & 0.241 & 56.08 & 0.087 & 63.14 & 0.150 \\
MWKR\_SPT & 34.81 & 0.051 & 30.50 & 0.086 & 20.12 & 0.150 & 17.65 & 0.419 & 35.43 & 0.243 & 29.47 & 0.084 & 39.94 & 0.146 \\
MWKR\_EET & 32.03 & 0.050 & 26.93 & 0.085 & 15.38 & 0.148 & 17.48 & 0.414 & 31.04 & 0.240 & 41.29 & 0.776 & 57.71 & 1.163 \\
\bottomrule
\end{tabular}
}
\end{table*}

\begin{table*}[ht!]
\centering
\caption{Optimality gap of ablated models. The best results are shown in bold. The relative optimality gap (Gap) is in percentage (\%).}
\label{tab: res: ablation}
\begin{tabular}{lrrrrrrrrrrrrrr}
\toprule
Method & \multicolumn{5}{c}{Low flexibility} & \multicolumn{2}{c}{High flexibility} \\
\cmidrule(lr){2-6} \cmidrule(lr){7-8} \\
 & \multicolumn{1}{c}{car8} & \multicolumn{1}{c}{mt10} & \multicolumn{1}{c}{la22} & \multicolumn{1}{c}{la32} & \multicolumn{1}{c}{la40} & \multicolumn{1}{c}{mt10} & \multicolumn{1}{c}{la22} \\
  & \multicolumn{1}{c}{8$\times$8} & \multicolumn{1}{c}{10$\times$10} & \multicolumn{1}{c}{15$\times$10} & \multicolumn{1}{c}{30$\times$10} & \multicolumn{1}{c}{15$\times$15} & \multicolumn{1}{c}{10$\times$10} & \multicolumn{1}{c}{15$\times$10} \\
 \midrule
2SL-FJSP & 1.47 &   \textbf{2.70} &   5.40 &   \textbf{9.50} & \   \textbf{3.16} &   \textbf{1.84} &   \textbf{12.38} \\
2SL-FJSP (\st{Branch}) & 1.64 & 3.36 & 6.39 & 10.49 & 3.39 & 1.91 & 19.44 \\
2SL-FJSP (\st{CR}) & 1.80 &   4.32 &   \textbf{5.11} &   10.58 & \   3.35 &   1.84 &   12.53  \\
2SL-FJSP (\st{SB}) & 8.04 &   6.32 &   15.40 &   15.23 & \   6.83 &   5.23 &   18.23  \\
2SL-FJSP (\st{SB},\st{CR}) & 8.87 &   7.52 &   16.23 &   16.43 & \   7.21 &   5.43 &   19.81  \\
Encoder & \textbf{0.26} &   6.04 &   15.48 &   12.71 & \   12.68 &   7.29 &   28.11  \\
Encoder (\st{SB}) & 4.70 &   10.21 &   16.88 &   15.32 & \   12.06 &   18.23 &   38.12  \\
\bottomrule
\end{tabular}
\end{table*}

\paragraph{Data Generation and Model Details}
A single instance of the FJSP contains $J$ jobs and $M$ machines with
a set of compatible assignments and the corresponding processing times
for each task.  Following \cite{kotary2022fast}, the instance
generation simulates a situation when a scheduling system experiences
an unexpected ``slowdown'' on some arbitrary machine, resulting in
increased processing times of each task assigned to the impacted
machine.  To construct the dataset, a base instance is selected from
the public benchmark \cite{hurink1994tabu} and a set of 20,000
individual problem instances are generated accordingly, with all the
processing times on the impaired machine increased by up to 50
percent.  Each FJSP instance is solved using IBM CP-Optimizer
solver \cite{laborie2009ibm} with a time limit of 1,800 seconds.  For
the JSP instances generated from the proposed branching policy, a
maximum of 20,000 instances are randomly selected and solved to train
the scheduling policy. The reference point for the symmetry-breaking
scheme is the solution of the instance with 25\% delay. The dataset is
split in 8:1:1 for training, validation, and testing. All
hyperparameter tunings are based on the validation metrics and all
results are reported on the testing set. The deep learning models are
implemented using PyTorch \cite{imambi2021pytorch} and the training
using Tesla V100 GPU on machines with Intel Xeon 2.7GHz and the Adam
optimizer \cite{kingma2014adam}. The hyperparameters were tuned using a
grid search and their setting are detailed in Appendix D.

\paragraph{Baseline Methods}

\tlf{} is compared with well-known rule-based heuristics from the literature. Like \tlf{}, the heuristics are
designed following a \textit{two-stage} hierarchical scheme. The first
stage involves a machine assignment rule, and the second stage
involves a job sequencing rule. Many such rules have been proposed in
the literature \cite{jun2019learning}. The experiments focus on eight
heuristics that were shown to be effective
(see \cite{jun2019learning,lei2022multi}). For \textit{machine
selection}, the experiments use the \textit{Smallest Processing Time}
(SPT) rule that assigns a machine with the smallest processing time of
a task and the \textit{Earliest End Time} (EET) rule that assigns a
machine with the earliest end time of a task. For \textit{job
sequencing}, the experiments use the \textit{First in First out}
(FIFO) rule that processes a job that arrives the earliest at the
queue of a machine, the \textit{Most Operation Number Remaining}
(MOPNR) rule that processes a job with the most number of tasks
remaining, the \textit{Least Work Remaining} (LWKR) rule that
processes a job with the least work remaining, and the \textit{Most
Work Remaining} (WMKR) rule that processes a job with the most work
remaining.

The experiments also compare \tlf{} against the state-of-the-art CP
commercial solver IBM CP-Optimizer \cite{laborie2009ibm} and the
state-of-the-art DRL method for
FJSP \cite{song2022flexible}\footnote{https://github.com/songwenas12/fjsp-drl}.
The DRL model is trained following the protocols defined
in \cite{song2022flexible}.

In addition, \tlf{} is compared with weaker versions of itself in an
ablation study:
\begin{itemize}
    \item Encoder: a one-stage deep learning approach following the parameterization in Section \ref{sec: architecture} trained on the symmetry-breaking dataset to predict the assignment and start time,
    \item Encoder (\st{SB}): a one-stage deep learning approach trained on the dataset from standard data generation;
    \item   2SL-FJSP (\st{Branch}): \tlf{} with no confidence-aware branching;
    \item   2SL-FJSP (\st{SB}): \tlf{} trained on the dataset from standard data generation;
    \item   2SL-FJSP (\st{CR}): \tlf{} trained on the dataset from symmetry-breaking data generation without constraint regularization;
    \item   2SL-FJSP (\st{SB}, \st{CR}): \tlf{} trained on the dataset from standard data generation without constraint regularization.
\end{itemize}

\subsection{Model Performance}

\paragraph{Optimality Gap}

The quality of the predictions is measured as a 1-shifted geometric
mean of the relative difference between the makespan of
the \textit{feasible solutions} recovered from the predictions of the
deep-learning models, and the makespan from the IBM CP solver with a
time limit of 1,800s i.e., \(\text{gap} =
(\text{makespan}_{\text{model}}
- \text{makespan}_{\text{cp}})/\text{makespan}_{\text{cp}}\).  The
optimality gap is the measure of the primary concern as the goal is to
predict solutions of FJSP as close to the optimal as possible.

Table \ref{tab: res: low_flexibility} reports the performance of the
different models.  First, \tlf{} significantly outperforms all other
baseline heuristics and the DRL method on all instances.  The gap of
\tlf{} is up to an order of magnitude smaller than the results of
the best heuristics on both low and high-flexibility instances.
Similarly, \tlf{} outperforms the one-stage deep learning approach
except on \texttt{car8}, the smallest instance.  This indicates that
explicitly modeling the structure of FJSP facilitates learning.
Interestingly, the DRL does not perform well on this set of test
cases. A possible reason is that the DRL is trained on
the dataset with a dramatic perturbation of jobs and machines
in \cite{song2022flexible}. This creates a distribution shift between the
training environments and the test environments.

\paragraph{Solving Time}

The solving times of different models are also reported in
Table \ref{tab: res: low_flexibility}. For \tlf{} and \texttt{Encoder},
the solving time includes the inference time and the solving time of
the feasibility recovery. The feasibility recovery is modeled as a
Linear Program (LP) and solved using Gurobi. In the experiments, all
feasibility recovery finishes in milliseconds since the LPs are all
solved in the presolving phase.  \tlf{} and \texttt{Encoder} are
orders of magnitude faster than the CP solver and DRL, and even one
order of magnitude faster than the fastest heuristics.

\paragraph{Ablation Study}

To understand the benefits of the proposed components in \tlf{},
Table \ref{tab: res: ablation} reports the results of weaker models
with different components ablated.  First, \tlf{} models are always
better than their one-stage counterparts \texttt{Encoder} except for
the smallest instance
\texttt{car8}. It indicates that explicitly modeling the hierarchical decisions
helps the learning model better capture the structure of FJSP.
Second, symmetry-breaking data generation shows great benefits for ML
models. It significantly reduces the optimality gap of all instances,
leading to up to 5 times smaller optimality gap for \tlf{} and 20
times smaller optimality gap for \texttt{Encoder}.  Finally, the
confidence-aware branching and constraint regularization help reduce
the optimality gap consistently for all instances except for the low
flexibility \texttt{la22}.

\section{Conclusion}

This paper proposes 2SL-FJSP, a two-stage learning framework that delivers, in milliseconds, high quality solution for FJSP. It uses a confidence-aware branching scheme that enables generating appropriate instances to train scheduling decisions based on assignment predictions. The paper proposes a novel symmetry breaking formulation of FJSP to improve the performance of the learning task. This formulation allows solving instances in parallel contrary to previous work in this area that is primarily sequential in nature. Computational results show that 2SL-FJSP is capable of generating feasible FJSP solutions with quality that is significantly better than other approaches used in practice. This work is a first step towards enabling effective real-time decision making in the context of FJSP environments, especially those characterized with low and high flexibility levels. Future work includes improving the learning framework to accommodate, effectively, more complex settings such as fully-flexible job shop problems, where each task can be processed on any machine.

\bibliographystyle{named}
\bibliography{ijcai22}

\begin{thebibliography}{}

\bibitem[\protect\citeauthoryear{Agrawal \bgroup \em et al.\egroup
  }{2019}]{agrawal2019differentiable}
Akshay Agrawal, Brandon Amos, Shane Barratt, Stephen Boyd, Steven Diamond, and
  J~Zico Kolter.
\newblock Differentiable convex optimization layers.
\newblock {\em Advances in neural information processing systems}, 32, 2019.

\bibitem[\protect\citeauthoryear{Amos and Kolter}{2017}]{amos2017optnet}
Brandon Amos and J~Zico Kolter.
\newblock Optnet: Differentiable optimization as a layer in neural networks.
\newblock In {\em International Conference on Machine Learning}, pages
  136--145. PMLR, 2017.

\bibitem[\protect\citeauthoryear{Bengio \bgroup \em et al.\egroup
  }{2021}]{bengio2021machine}
Yoshua Bengio, Andrea Lodi, and Antoine Prouvost.
\newblock Machine learning for combinatorial optimization: a methodological
  tour d’horizon.
\newblock {\em European Journal of Operational Research}, 290(2):405--421,
  2021.

\bibitem[\protect\citeauthoryear{Brandimarte}{1993}]{brandimarte1993routing}
Paolo Brandimarte.
\newblock Routing and scheduling in a flexible job shop by tabu search.
\newblock {\em Annals of Operations research}, 41(3):157--183, 1993.

\bibitem[\protect\citeauthoryear{Chatzos \bgroup \em et al.\egroup
  }{2020}]{chatzos2020high}
Minas Chatzos, Ferdinando Fioretto, Terrence~WK Mak, and Pascal Van~Hentenryck.
\newblock High-fidelity machine learning approximations of large-scale optimal
  power flow.
\newblock {\em arXiv preprint arXiv:2006.16356}, 2020.

\bibitem[\protect\citeauthoryear{Chaudhry and
  Khan}{2016}]{chaudhry2016research}
Imran~Ali Chaudhry and Abid~Ali Khan.
\newblock A research survey: review of flexible job shop scheduling techniques.
\newblock {\em International Transactions in Operational Research},
  23(3):551--591, 2016.

\bibitem[\protect\citeauthoryear{Chen \bgroup \em et al.\egroup
  }{2020}]{chen2020rna}
Xinshi Chen, Yu~Li, Ramzan Umarov, Xin Gao, and Le~Song.
\newblock Rna secondary structure prediction by learning unrolled algorithms.
\newblock {\em arXiv preprint arXiv:2002.05810}, 2020.

\bibitem[\protect\citeauthoryear{Donti \bgroup \em et al.\egroup
  }{2021}]{donti2021dc3}
Priya~L Donti, David Rolnick, and J~Zico Kolter.
\newblock Dc3: A learning method for optimization with hard constraints.
\newblock {\em arXiv preprint arXiv:2104.12225}, 2021.

\bibitem[\protect\citeauthoryear{Fattahi \bgroup \em et al.\egroup
  }{2007}]{fattahi2007mathematical}
Parviz Fattahi, Mohammad Saidi~Mehrabad, and Fariborz Jolai.
\newblock Mathematical modeling and heuristic approaches to flexible job shop
  scheduling problems.
\newblock {\em Journal of intelligent manufacturing}, 18(3):331--342, 2007.

\bibitem[\protect\citeauthoryear{Fioretto \bgroup \em et al.\egroup
  }{2020}]{fioretto2020predicting}
Ferdinando Fioretto, Terrence~WK Mak, and Pascal Van~Hentenryck.
\newblock Predicting ac optimal power flows: Combining deep learning and
  lagrangian dual methods.
\newblock In {\em Proceedings of the AAAI Conference on Artificial
  Intelligence}, volume~34, pages 630--637, 2020.

\bibitem[\protect\citeauthoryear{Gal and Ghahramani}{2016}]{gal2016dropout}
Yarin Gal and Zoubin Ghahramani.
\newblock Dropout as a bayesian approximation: Representing model uncertainty
  in deep learning.
\newblock In {\em international conference on machine learning}, pages
  1050--1059. PMLR, 2016.

\bibitem[\protect\citeauthoryear{Hurink \bgroup \em et al.\egroup
  }{1994}]{hurink1994tabu}
Johann Hurink, Bernd Jurisch, and Monika Thole.
\newblock Tabu search for the job-shop scheduling problem with multi-purpose
  machines.
\newblock {\em Operations-Research-Spektrum}, 15(4):205--215, 1994.

\bibitem[\protect\citeauthoryear{Imambi \bgroup \em et al.\egroup
  }{2021}]{imambi2021pytorch}
Sagar Imambi, Kolla~Bhanu Prakash, and GR~Kanagachidambaresan.
\newblock Pytorch.
\newblock In {\em Programming with TensorFlow}, pages 87--104. Springer, 2021.

\bibitem[\protect\citeauthoryear{Jun \bgroup \em et al.\egroup
  }{2019}]{jun2019learning}
Sungbum Jun, Seokcheon Lee, and Hyonho Chun.
\newblock Learning dispatching rules using random forest in flexible job shop
  scheduling problems.
\newblock {\em International Journal of Production Research},
  57(10):3290--3310, 2019.

\bibitem[\protect\citeauthoryear{Kawai and Fujimoto}{2005}]{kawai2005efficient}
Tatsunobu Kawai and Yasutaka Fujimoto.
\newblock An efficient combination of dispatch rules for job-shop scheduling
  problem.
\newblock In {\em INDIN'05. 2005 3rd IEEE International Conference on
  Industrial Informatics, 2005.}, pages 484--488. IEEE, 2005.

\bibitem[\protect\citeauthoryear{Khalil \bgroup \em et al.\egroup
  }{2017}]{khalil2017learning}
Elias Khalil, Hanjun Dai, Yuyu Zhang, Bistra Dilkina, and Le~Song.
\newblock Learning combinatorial optimization algorithms over graphs.
\newblock {\em Advances in neural information processing systems}, 30, 2017.

\bibitem[\protect\citeauthoryear{Kingma and Ba}{2014}]{kingma2014adam}
Diederik~P Kingma and Jimmy Ba.
\newblock Adam: A method for stochastic optimization.
\newblock {\em arXiv preprint arXiv:1412.6980}, 2014.

\bibitem[\protect\citeauthoryear{Kool \bgroup \em et al.\egroup
  }{2018}]{kool2018attention}
Wouter Kool, Herke Van~Hoof, and Max Welling.
\newblock Attention, learn to solve routing problems!
\newblock {\em arXiv preprint arXiv:1803.08475}, 2018.

\bibitem[\protect\citeauthoryear{Kotary \bgroup \em et al.\egroup
  }{2021a}]{kotary2021learning}
James Kotary, Ferdinando Fioretto, and Pascal Van~Hentenryck.
\newblock Learning hard optimization problems: A data generation perspective.
\newblock {\em Advances in Neural Information Processing Systems},
  34:24981--24992, 2021.

\bibitem[\protect\citeauthoryear{Kotary \bgroup \em et al.\egroup
  }{2021b}]{kotary2021end}
James Kotary, Ferdinando Fioretto, Pascal Van~Hentenryck, and Bryan Wilder.
\newblock End-to-end constrained optimization learning: A survey.
\newblock {\em arXiv preprint arXiv:2103.16378}, 2021.

\bibitem[\protect\citeauthoryear{Kotary \bgroup \em et al.\egroup
  }{2022}]{kotary2022fast}
James Kotary, Ferdinando Fioretto, and Pascal Van~Hentenryck.
\newblock Fast approximations for job shop scheduling: A lagrangian dual deep
  learning method.
\newblock In {\em Proceedings of the AAAI Conference on Artificial
  Intelligence}, volume~36, pages 7239--7246, 2022.

\bibitem[\protect\citeauthoryear{Laborie}{2009}]{laborie2009ibm}
Philippe Laborie.
\newblock Ibm ilog cp optimizer for detailed scheduling illustrated on three
  problems.
\newblock In {\em International Conference on Integration of Constraint
  Programming, Artificial Intelligence, and Operations Research}, pages
  148--162. Springer, 2009.

\bibitem[\protect\citeauthoryear{Lei \bgroup \em et al.\egroup
  }{2022}]{lei2022multi}
Kun Lei, Peng Guo, Wenchao Zhao, Yi~Wang, Linmao Qian, Xiangyin Meng, and
  Liansheng Tang.
\newblock A multi-action deep reinforcement learning framework for flexible
  job-shop scheduling problem.
\newblock {\em Expert Systems with Applications}, page 117796, 2022.

\bibitem[\protect\citeauthoryear{Lin \bgroup \em et al.\egroup
  }{2019}]{lin2019smart}
Chun-Cheng Lin, Der-Jiunn Deng, Yen-Ling Chih, and Hsin-Ting Chiu.
\newblock Smart manufacturing scheduling with edge computing using multiclass
  deep q network.
\newblock {\em IEEE Transactions on Industrial Informatics}, 15(7):4276--4284,
  2019.

\bibitem[\protect\citeauthoryear{Liu \bgroup \em et al.\egroup
  }{2022}]{liu2022deep}
Renke Liu, Rajesh Piplani, and Carlos Toro.
\newblock Deep reinforcement learning for dynamic scheduling of a flexible job
  shop.
\newblock {\em International Journal of Production Research}, pages 1--21,
  2022.

\bibitem[\protect\citeauthoryear{Naderi and
  Roshanaei}{2022}]{naderi2022critical}
Bahman Naderi and Vahid Roshanaei.
\newblock Critical-path-search logic-based benders decomposition approaches for
  flexible job shop scheduling.
\newblock {\em INFORMS Journal on Optimization}, 4(1):1--28, 2022.

\bibitem[\protect\citeauthoryear{Ostrowski \bgroup \em et al.\egroup
  }{2010}]{ostrowski2010symmetry}
James Ostrowski, Miguel~F Anjos, and Anthony Vannelli.
\newblock {\em Symmetry in scheduling problems}.
\newblock Citeseer, 2010.

\bibitem[\protect\citeauthoryear{Park \bgroup \em et al.\egroup
  }{2019}]{park2019reinforcement}
In-Beom Park, Jaeseok Huh, Joongkyun Kim, and Jonghun Park.
\newblock A reinforcement learning approach to robust scheduling of
  semiconductor manufacturing facilities.
\newblock {\em IEEE Transactions on Automation Science and Engineering},
  17(3):1420--1431, 2019.

\bibitem[\protect\citeauthoryear{Park \bgroup \em et al.\egroup
  }{2022}]{park2022confidence}
Seonho Park, Wenbo Chen, Dahye Han, Mathieu Tanneau, and Pascal Van~Hentenryck.
\newblock Confidence-aware graph neural networks for learning reliability
  assessment commitments.
\newblock {\em arXiv preprint arXiv:2211.15755}, 2022.

\bibitem[\protect\citeauthoryear{Rahaman and
  others}{2021}]{rahaman2021uncertainty}
Rahul Rahaman et~al.
\newblock Uncertainty quantification and deep ensembles.
\newblock {\em Advances in Neural Information Processing Systems},
  34:20063--20075, 2021.

\bibitem[\protect\citeauthoryear{Schulman \bgroup \em et al.\egroup
  }{2017}]{schulman2017proximal}
John Schulman, Filip Wolski, Prafulla Dhariwal, Alec Radford, and Oleg Klimov.
\newblock Proximal policy optimization algorithms.
\newblock {\em arXiv preprint arXiv:1707.06347}, 2017.

\bibitem[\protect\citeauthoryear{Song \bgroup \em et al.\egroup
  }{2022}]{song2022flexible}
Wen Song, Xinyang Chen, Qiqiang Li, and Zhiguang Cao.
\newblock Flexible job shop scheduling via graph neural network and deep
  reinforcement learning.
\newblock {\em IEEE Transactions on Industrial Informatics}, 2022.

\bibitem[\protect\citeauthoryear{Vesselinova \bgroup \em et al.\egroup
  }{2020}]{vesselinova2020learning}
Natalia Vesselinova, Rebecca Steinert, Daniel~F Perez-Ramirez, and Magnus
  Boman.
\newblock Learning combinatorial optimization on graphs: A survey with
  applications to networking.
\newblock {\em IEEE Access}, 8:120388--120416, 2020.

\bibitem[\protect\citeauthoryear{Vinyals \bgroup \em et al.\egroup
  }{2015}]{vinyals2015pointer}
Oriol Vinyals, Meire Fortunato, and Navdeep Jaitly.
\newblock Pointer networks.
\newblock {\em Advances in neural information processing systems}, 28, 2015.

\bibitem[\protect\citeauthoryear{Vlastelica \bgroup \em et al.\egroup
  }{2019}]{vlastelica2019differentiation}
Marin Vlastelica, Anselm Paulus, V{\'\i}t Musil, Georg Martius, and Michal
  Rol{\'\i}nek.
\newblock Differentiation of blackbox combinatorial solvers.
\newblock {\em arXiv preprint arXiv:1912.02175}, 2019.

\bibitem[\protect\citeauthoryear{Xie \bgroup \em et al.\egroup
  }{2019}]{xie2019review}
Jin Xie, Liang Gao, Kunkun Peng, Xinyu Li, and Haoran Li.
\newblock Review on flexible job shop scheduling.
\newblock {\em IET Collaborative Intelligent Manufacturing}, 1(3):67--77, 2019.

\bibitem[\protect\citeauthoryear{Zhang \bgroup \em et al.\egroup
  }{2020}]{zhang2020learning}
Cong Zhang, Wen Song, Zhiguang Cao, Jie Zhang, Puay~Siew Tan, and Xu~Chi.
\newblock Learning to dispatch for job shop scheduling via deep reinforcement
  learning.
\newblock {\em Advances in Neural Information Processing Systems},
  33:1621--1632, 2020.

\end{thebibliography}

\clearpage
\appendix
\section{FJSP CP formulation}
\label{appendix: detailed_cp_formulation}

A Constraint Programming (CP) formulation is proposed and used for solving FJSP. Two interval variables are used, $ops$ denoting a decision variable for each job $j$ and task $t$, and $mops$ denoting an optional decision variable for each job $j$ and its task $t$ on compatible machine $m$. Each interval variable has a domain $(s,e)$, where $s$ and $e$ are the start and end times of the interval, respectively, and $d=s-e$ is its length, in time units. The CP formulation is shown in Model 4 below. The objective (\ref{eq:fjsp_obj_simple_CP}) minimizes the makespan. Constraints \ref{fjsp_machineSelection_CP}, (\ref{fjsp_jobPrecedence_CP}), and \ref{fjsp_noOverlap_CP} are the assignment, job precedence, and machine precedence constraints, respectively.

\begin{model}[ht!]
\begin{subequations}
{\scriptsize
\caption{A CP formulation for the FJSP}
\label{model:fjsp_cp}
\begin{flalign} 
&\code{Minimize} \quad \code{max(end\_of(ops[j,t]) for (j,t) in ops)} &\label{eq:fjsp_obj_simple_CP}\\
&\code{subject to:}\nonumber\\
\begin{split}
\code{alternative(ops[j,t],[mops[j,t,m]:(j,t,m)} \in \code{mops]} \\
\forall \code{(j,t)} \in \code{ops}, 
\end{split} \label{fjsp_machineSelection_CP}\\
\begin{split}
\code{end\_before\_start(ops[j,t],ops[j,t-1])} \\ \forall \code{(j,t)} \in \code{ops,t>0}, \end{split} \label{fjsp_jobPrecedence_CP}\\ 
\begin{split} \code{no\_overlap(mops[j,t,m] for (j,t,m)}  \\ \forall \code{m} \in \code{M}.
\end{split}\label{fjsp_noOverlap_CP}
\end{flalign}
}
\end{subequations}
\end{model}

\section{Feasibility Recovery} \label{appendix: feasibility_recovery}

To recover feasibility, 2SL-FJSP utilizes a polynomial time procedure proposed in \cite{kotary2022fast}. It relies on the fact that a predicted start time vector $\hat{s}$ implies an implicit task ordering between tasks assigned to the same machine
Model 5 represents a linear programming (LP) formulation that finds the optimal schedule subject to such ordering. The objective function (\ref{eq:feasibility_recovery_obj}) along with the job precedence constraints (\ref{eq:fr_jobPrec}) are identical to those of Model 2. Constraints (\ref{eq:fr_machinePrec}) reflects machine precedence constraints given task ordering, thus, eliminating the need for a disjunctive constraint set similar to that of Model 2. Using Model 5 was sufficient to restore feasibility in all of our experiments. However, it may happen that the resulting LP is infeasible as the machine precedence constraints may not be always consistent with the job precedence constraints. To address this, a greedy recovery algorithm can be used, as proposed in \cite{kotary2022fast}.

\begin{model}[ht!]
\begin{subequations}
{\scriptsize
\caption{LP formulation for feasibility recovery}
\label{model:feasibilityRecovery_LP}
\begin{flalign}
&\Minimize_{}  \quad C_{max} & ,\label{eq:feasibility_recovery_obj}\\
&\text{\textbf{subject to:}}\nonumber\\
&  \qquad s_{jt-1} \geq  s_{jt} + d_{jt} \;  & \forall j \in [J], t \in [T], t>0, \label{eq:fr_jobPrec}\\
 & \qquad s_{jt} \geq  s_{j^\prime t^\prime} + d_{j^\prime t^\prime} & \forall j, j^\prime \in [J], t, t^\prime \in [T]: (j^\prime, t^\prime) \prec_{\hat{s}} (j, t), \label{eq:fr_machinePrec} \\
& \qquad C_{max} \geq  s_{jt} + d_{jt} \; & \forall j \in [J], t \in [T], t=0, \\
 & \qquad  s_{jt} \geq 0 & \forall j \in [J], t \in [T].
\end{flalign}
}
\end{subequations}
\end{model}

\section{FJSP CP Symmetry Breaking formulation}
\label{appendix: detailed_cp_SB_formulation}

A Constraint Programming (CP) formulation is proposed and used for solving the \textit{Symmetry Breaking} version of FJSP. The description of decision variables and constraints is identical to that of the standard FJSP presented in Appendix A. The objective function (\ref{obj_SB_fjsp}) is modified to minimize the deviation of the optimal solution from a reference solution. Constraints (\ref{fjsp_SB_makespan_CP}) ensures that the quality of the
symmetry-breaking solution is not worse than the optimal solution to
the original FJSP.

\begin{model}[ht!]
\begin{subequations}
{\scriptsize
\caption{A Symmetry Breaking CP formulation for the FJSP}
\label{model:fjsp_cp_SB}
\begin{flalign} 
\begin{split}
\code{Minimize} \qquad \code{sum(abs(presence\_of(mops[j,t,m]) - 1)} \\ \code{for (j,t,m)} \in \code{ref\_plan )}
 \end{split} \label{obj_SB_fjsp}
\\
&\code{subject to:}\nonumber\\
\begin{split}
\code{end\_before\_start(ops[j,t],ops[j,t-1])} \\ \forall \code{(j,t)} \in \code{ops,t>0}, \end{split} \label{fjsp_SB_jobPrecedence_CP}\\ 
\begin{split}
\code{alternative(ops[j,t],[mops[j,t,m]:(j,t,m)} \in \code{mops]} \\
\forall \code{(j,t)} \in \code{ops}, 
\end{split} \label{fjsp_SB_machineSelection_CP}\\
\begin{split} \code{no\_overlap(mops[j,t,m] for (j,t,m)}  \\ \forall \code{m} \in \code{M},
\end{split}\label{fjsp_SB_noOverlap_CP} \\
& \code{C}^*_{max} \geq \code{max(end\_of(ops[j,t]) for (j,t) in ops)}  \label{fjsp_SB_makespan_CP}
\end{flalign}
}
\end{subequations}
\end{model}

\section{Hyperparameters Tuning of Deep Learning Models} 
The following parameters were tuned given the following range for all neural network-based methods across all experiments:
\begin{itemize}
    \item Maximum epochs: 1000
    \item Batch size: 32
    \item Encoding Layers: 2, 3
    \item Decoding Layers: 2, 3
    \item Dropout: 0.2
    \item Learning rate: 1e-1, 1e-2, 1e-3
\end{itemize}
\label{appendix: tuning}
In training, the learning rate is reduced to a factor of 10 if the validation metric does not improve for consecutive 10 epochs and the training early stops if the validation metric does not improve for consecutive 20 epochs.
The hyperparameters of the model are selected by the smallest validation loss.
The best hyperparameters of different models on different instances are reported in Table \ref{ref:hyperparameter:car8}, \ref{ref:hyperparameter:mt10}, \ref{ref:hyperparameter:la22},\ref{ref:hyperparameter:la32},\ref{ref:hyperparameter:mt10:high},\ref{ref:hyperparameter:la22:high}, respectively.

\begin{table}[h]
\caption{The selected hyperparameters for car8 instances}
\label{ref:hyperparameter:car8}
\resizebox{0.99\columnwidth}{!}{
\begin{tabular}{llrrr}
\toprule
dataset & model & \multicolumn{1}{l}{lr} & \multicolumn{1}{l}{Enc. Layer} & \multicolumn{1}{l}{Dec. Layer} \\
\midrule
SB & 2SL-FJSP (cls) & 0.01 & 2 & 2 \\
 & 2SL-FJSP (reg) & 0.1 & 3 & 3 \\
 & Encoder (cls) & 0.01 & 2 & 2 \\
 & Encoder (reg) & 0.1 & 3 & 3 \\
 \midrule
Standard & 2SL-FJSP (cls) & 0.1 & 2 & 2 \\
 & 2SL-FJSP (reg) & 0.01 & 2 & 2 \\
 & Encoder (cls) & 0.1 & 2 & 2 \\
 & Encoder (reg) & 0.01 & 2 & 2 \\
 \bottomrule
\end{tabular}
}
\end{table}

\begin{table}[]
\caption{The selected hyperparameters for mt10 instances}
\label{ref:hyperparameter:mt10}
\resizebox{0.99\columnwidth}{!}{
\begin{tabular}{llrrr}
\toprule
dataset & model & \multicolumn{1}{l}{lr} & \multicolumn{1}{l}{Enc. Layer} & \multicolumn{1}{l}{Dec. Layer} \\
\midrule
SB & 2SL-FJSP (cls) & 0.01 & 2 & 2 \\
 & 2SL-FJSP (reg) & 0.01 & 3 & 3 \\
 & Encoder (cls) & 0.01 & 2 & 2 \\
 & Encoder (reg) & 0.01 & 2 & 2 \\
 \midrule
Standard & 2SL-FJSP (cls) & 0.01 & 2 & 2 \\
 & 2SL-FJSP (reg) & 0.01 & 3 & 3 \\
 & Encoder (cls) & 0.01 & 2 & 2 \\
 & Encoder (reg) & 0.01 & 2 & 2 \\
 \bottomrule
\end{tabular}
}
\end{table}

\begin{table}[]
\caption{The selected hyperparameters for la22 instances}
\label{ref:hyperparameter:la22}
\resizebox{0.99\columnwidth}{!}{
\begin{tabular}{llrrr}
\toprule
dataset & model & \multicolumn{1}{l}{lr} & \multicolumn{1}{l}{Enc. Layer} & \multicolumn{1}{l}{Dec. Layer} \\
\midrule
SB & 2SL-FJSP (cls) & 0.01 & 2 & 2 \\
 & 2SL-FJSP (reg) & 0.001 & 2 & 2 \\
 & Encoder (cls) & 0.01 & 2 & 2 \\
 & Encoder (reg) & 0.1 & 3 & 3 \\
 \midrule
Standard & 2SL-FJSP (cls) & 0.01 & 3 & 3 \\
 & 2SL-FJSP (reg) & 0.1 & 3 & 3 \\
 & Encoder (cls) & 0.01 & 3 & 3 \\
 & Encoder (reg) & 0.1 & 2 & 2 \\
 \bottomrule
\end{tabular}
}
\end{table}

\begin{table}[]
\caption{The selected hyperparameters for la32 instances}
\label{ref:hyperparameter:la32}
\resizebox{0.99\columnwidth}{!}{
\begin{tabular}{llrrr}
\toprule
dataset & model & \multicolumn{1}{l}{lr} & \multicolumn{1}{l}{Enc. Layer} & \multicolumn{1}{l}{Dec. Layer} \\
\midrule
SB & 2SL-FJSP (cls) & 0.01 & 3 & 3 \\
 & 2SL-FJSP (reg) & 0.01 & 3 & 3 \\
 & Encoder (cls) & 0.01 & 3 & 3 \\
 & Encoder (reg) & 0.1 & 3 & 3 \\
 \midrule
Standard & 2SL-FJSP (cls) & 0.01 & 3 & 3 \\
 & 2SL-FJSP (reg) & 0.01 & 3 & 3 \\
 & Encoder (cls) & 0.01 & 2 & 2 \\
 & Encoder (reg) & 0.1 & 3 & 3 \\
 \bottomrule
\end{tabular}
}
\end{table}

\begin{table}[]
\caption{The selected hyperparameters for high flexible mt10 instances}
\label{ref:hyperparameter:mt10:high}
\resizebox{0.99\columnwidth}{!}{
\begin{tabular}{llrrr}
\toprule
dataset & model & \multicolumn{1}{l}{lr} & \multicolumn{1}{l}{Enc. Layer} & \multicolumn{1}{l}{Dec. Layer} \\
\midrule
SB & 2SL-FJSP (cls) & 0.1 & 3 & 3 \\
 & 2SL-FJSP (reg) & 0.1 & 2 & 2 \\
 & Encoder (cls) & 0.1 & 3 & 3 \\
 & Encoder (reg) & 0.1 & 2 & 2 \\
\midrule
Standard & 2SL-FJSP (cls) & 0.1 & 3 & 3 \\
 & 2SL-FJSP (reg) & 0.1 & 2 & 2 \\
 & Encoder (cls) & 0.1 & 3 & 3 \\
 & Encoder (reg) & 0.1 & 2 & 2 \\
 \bottomrule
\end{tabular}
}
\end{table}

\begin{table}[]
\caption{The selected hyperparameters for high flexible la22 instances}
\label{ref:hyperparameter:la22:high}
\resizebox{0.99\columnwidth}{!}{
\begin{tabular}{llrrr}
\toprule
dataset & model & \multicolumn{1}{l}{lr} & \multicolumn{1}{l}{Enc. Layer} & \multicolumn{1}{l}{Dec. Layer} \\
\midrule
SB & 2SL-FJSP (cls) & 0.1 & 3 & 3 \\
 & 2SL-FJSP (reg) & 0.01 & 3 & 3 \\
 & Encoder (cls) & 0.1 & 3 & 3 \\
 & Encoder (reg) & 0.01 & 2 & 2 \\
\midrule
Standard & 2SL-FJSP (cls) & 0.1 & 3 & 3 \\
 & 2SL-FJSP (reg) & 0.01 & 3 & 3 \\
 & Encoder (cls) & 0.1 & 3 & 3 \\
 & Encoder (reg) & 0.01 & 2 & 2 \\
 \bottomrule
\end{tabular}
}
\end{table}

\end{document}